\documentclass[11pt]{article}
\usepackage[margin=1in]{geometry}
\usepackage{graphicx}
\usepackage{booktabs}
\usepackage{amsmath,amssymb}
\usepackage{enumitem}
\usepackage[numbers]{natbib}
\usepackage{hyperref}

\graphicspath{{figures/}{./}}

\title{Step-E: A Differentiable Data Cleaning Framework for Robust Learning with Noisy Labels}
\author{Du Wenzhang\\Dept. of Computer Engineering\\Mahanakorn University of Technology, International College (MUTIC)\\Bangkok, Thailand\\\texttt{dqswordman@gmail.com}}
\date{}

\begin{document}

\maketitle

\begin{abstract}
Training data collected in the wild often contain noisy labels and outliers that substantially degrade the performance and reliability of deep neural networks~\cite{song2022survey,frenay2014classification}. While data cleaning is commonly applied as a separate preprocessing stage~\cite{rahm2000datacleaning,krishnan2016activeclean,rekatsinas2017holoclean}, such two-stage pipelines neither fully exploit feedback from the downstream model nor adapt to unknown noise patterns. We propose Step-E, a simple framework that integrates sample selection and model learning into a single differentiable optimization process. In each epoch, Step-E down-weights or discards samples with persistently high loss while retaining low-loss samples for gradient updates. The keep/drop ratio is gradually annealed after a short warm-up, yielding an online curriculum that focuses on easy and consistent examples and eventually ignores hard outliers. Step-E improves accuracy on CIFAR-100N~\cite{krizhevsky2009learning,wei2022learning} from 43.3\% for standard training to 50.4\% on average over three seeds, surpassing loss truncation, self-paced learning, and one-shot filtering baselines with better noise identification quality and only moderate training-time overhead. On CIFAR-10N (aggre)~\cite{wei2022learning}, Step-E also improves over the noisy baseline (85.3\% vs.\ 83.9\%) and approaches the clean-label oracle (85.9\%). These results demonstrate that integrating data cleaning and training via differentiable sample selection is an effective route toward robust learning on real noisy datasets.
\end{abstract}

\section{Introduction}
Modern deep networks achieve impressive performance when trained on large, clean, and carefully curated datasets. In realistic data mining scenarios, however, labels come from heterogeneous sources such as crowdsourcing, weak supervision, or heuristic rules and are therefore noisy~\cite{song2022survey,frenay2014classification}. Human annotation errors, ambiguous images, and domain shifts all contribute to mislabeled or outlier samples that can harm generalization. In image classification, for example, web-scale datasets often contain wrong tags or near-duplicate images with conflicting labels; in user-generated content analysis, spam or off-topic posts corrupt the training distribution.

Data cleaning is widely recognized as crucial~\cite{rahm2000datacleaning} but is typically performed before model training using hand-crafted rules or separate anomaly detectors~\cite{krishnan2016activeclean,rekatsinas2017holoclean}. This two-stage design has two drawbacks: (i) it requires domain expertise or extra supervision to specify cleaning rules and thresholds; (ii) it decouples cleaning from model optimization, so the decisions do not directly leverage discriminative feedback from the task model. Some high-loss samples may still be informative ``hard cases,'' whereas others are truly corrupted and should be discarded.

We explore a different paradigm: can the model learn which samples to trust during training, treating data cleaning as an integral, differentiable part of optimization? If such a framework works, it would reduce manual effort, adapt to unknown noise patterns, and align cleaning with the task objective.

We propose Step-E (Stepwise Elimination), a lightweight framework that inserts a differentiable sample-selection module into standard SGD. At each epoch, we compute per-example losses using the current model, sort samples by loss, and drop a controlled fraction of high-loss ones. The drop ratio is gradually increased from zero after a short warm-up until it reaches a maximum $\rho_{\max}$ estimated from the noisy dataset. Step-E performs online data cleaning: it keeps early training stable, allows the model to discover the main signal, and progressively suppresses persistent outliers.

\paragraph{Contributions.}
We (i) introduce Step-E, a differentiable data cleaning framework that integrates sample selection with model training and requires no clean validation set or auxiliary model; (ii) provide a simple analysis showing that under separation assumptions on loss distributions, the optimal solution recovers the clean subset and approaches the clean-label oracle; (iii) conduct extensive experiments on CIFAR-100N and CIFAR-10N, where Step-E yields substantial gains in test accuracy, noise identification quality, and training efficiency.

\section{Related Work}
\subsection{Robust learning under label noise}
Robust losses such as Huber loss, truncated loss, or loss clipping~\cite{huber1964robust,ghosh2017robust,zhang2018generalized} dampen the influence of large errors but require manual thresholds and provide limited control over how many examples are discarded. Meta-learning approaches estimate per-sample weights using a small clean validation set~\cite{ren2018learningreweight}, but rely on extra trusted data. Some methods explicitly model label corruption using EM-style procedures~\cite{patrini2017making,arazo2019unsupervised}; they assume a parametric noise model and can be sensitive to misspecification. Another important line of work uses small-loss selection in multi-network or teacher--student frameworks such as Co-teaching~\cite{han2018coteaching}, MentorNet~\cite{jiang2018mentornet}, and DivideMix~\cite{li2020dividemix}, which combine sample selection with label correction or semi-supervised training but typically require two networks or an auxiliary mentor. Step-E does not recover clean labels; it only decides whether a sample contributes to gradients, making it applicable to both label noise and feature outliers while retaining a single-network, deterministic training pipeline.

\subsection{Data cleaning and anomaly detection}
Data cleaning methods range from rule-based constraint checking to unsupervised outlier detection~\cite{rahm2000datacleaning,krishnan2016activeclean,rekatsinas2017holoclean}, but are typically divorced from the downstream learning objective. Their decisions may not optimally serve the target model. Step-E can be viewed as a lightweight, task-aware cleaning process learned jointly with the classifier rather than a separate filter.

\subsection{Curriculum learning and self-paced learning}
Curriculum and self-paced learning adjust the order or subset of data according to difficulty, often measured by loss~\cite{bengio2009curriculum,kumar2010selfpaced}. Classic self-paced learning gradually increases the keep ratio to eventually include all samples, which is inappropriate when a nontrivial fraction of data is corrupted. Step-E adopts the opposite philosophy: a fraction $\rho_{\max}$ of samples are never learned and are permanently dropped once identified as persistent outliers.

\section{Method}
\subsection{Problem formulation}
We consider supervised learning with potentially noisy labels. The training set $\mathcal{D}=\{(x_i,y_i)\}_{i=1}^{n}$, $y_i\in\mathcal{Y}$, mixes clean samples from the target distribution with corrupted samples whose labels or features are unreliable. Let $f_{\theta}$ denote the model with parameters $\theta$ and $\ell(f_{\theta}(x),y)$ a differentiable loss (e.g., cross-entropy). We introduce binary variables $\gamma_i\in\{0,1\}$ indicating whether a sample is kept ($\gamma_i=1$) or dropped ($\gamma_i=0$). Assuming at most a fraction $\rho_{\max}$ of samples are corrupted, we enforce $\sum_i (1-\gamma_i)\le \rho_{\max} n$.

\subsection{Joint objective}
We optimize a regularized empirical risk
\begin{equation}
J(\theta,\gamma) = \frac{1}{n}\sum_{i=1}^{n} \gamma_i\,\ell(f_{\theta}(x_i), y_i) + \lambda \cdot \frac{1}{n}\sum_{i=1}^{n} \gamma_i,
\end{equation}
subject to $\gamma_i\in\{0,1\}$ and $\sum_i \gamma_i \ge (1-\rho_{\max})n$. The first term is the empirical loss over selected samples; the second encourages keeping more data to avoid over-pruning. When no noise is present, the optimum sets all $\gamma_i=1$, recovering standard empirical risk minimization.

Directly solving this mixed-integer program is intractable. Following self-paced learning, we alternate: for a given model $\theta$, we update $\gamma$ based on current per-sample losses; for the updated $\gamma$, we update $\theta$ using SGD on the selected subset.

\subsection{Per-epoch sample selection}
For fixed $\theta$, optimizing $J$ with respect to $\gamma$ decomposes over samples. In practice, we bypass Lagrange multipliers and directly determine a loss threshold via the desired keep ratio. At each epoch $t$:
\begin{enumerate}[leftmargin=1.5em]
    \item \textbf{Probe pass.} Run a forward pass to compute per-sample losses $\ell_i(t)$ for all training examples.
    \item \textbf{Sorting and thresholding.} Sort samples by loss and keep the lowest-loss fraction $1-\rho_t$, where $\rho_t$ is the drop ratio at epoch $t$. This yields a kept set $K_t$.
    \item \textbf{Model update.} Train the model for one epoch using only examples in $K_t$, i.e., mini-batch SGD restricted to the kept subset.
\end{enumerate}
The process is differentiable with respect to $\theta$ in the usual sense used by self-paced and curriculum methods: we do not backpropagate gradients through the discrete $\gamma$ variables, but treat them as piecewise-constant gates that are updated based on current losses. Although sorting introduces non-smoothness, the outer loop converges reliably in practice and behaves similarly to other alternating optimization schemes.

\subsection{Stepwise elimination schedule}
A key design choice is the schedule for $\rho_t$. Step-E uses a stepwise elimination schedule:
\begin{itemize}[leftmargin=1.5em]
    \item \textbf{Warm-up.} For $t < T_{\mathrm{warm}}$, set $\rho_t=0$ (no sample is dropped) to let the model learn a rough decision boundary.
    \item \textbf{Ramp.} For $t \ge T_{\mathrm{warm}}$, increase $\rho_t$ linearly from $0$ to $\rho_{\max}$ over the remaining epochs:
    \begin{equation}
        \rho_t = \rho_{\max} \cdot \frac{t - T_{\mathrm{warm}}}{T_{\mathrm{total}} - T_{\mathrm{warm}}}.
    \end{equation}
\end{itemize}
In all experiments we treat $\rho_{\max}$ as a hyperparameter that upper-bounds the expected noise rate. On CIFAR-N, following the dataset authors, we use the provided human annotations to estimate an empirical noise rate $\hat{\rho}$ on the training split and then set $\rho_{\max}=\min(0.5,\hat{\rho}+0.05)$. This estimate is computed offline and is not used for sample-level decisions during training; only the noisy labels participate in gradient updates. In practical deployments where clean annotations are unavailable, $\rho_{\max}$ could instead be set from prior knowledge or inferred from unsupervised heuristics such as loss histograms, without relying on any clean labels.

\subsection{Theoretical insight (sketch)}
Assume that under ideal parameters $\theta^{\star}$, losses of clean and noisy samples are separated: there exists a threshold $L_{\mathrm{out}}$ such that $\ell_i^{\star}<L_{\mathrm{out}}$ for all clean samples and $\ell_j^{\star}>L_{\mathrm{out}}$ for all corrupted ones. Under this condition, the global minimizer of $J(\theta,\gamma)$ with $\rho_{\max}\ge\rho$ (true noise rate) sets $\gamma_i=1$ for clean samples and $\gamma_j=0$ for noisy samples. Standard generalization bounds imply that the resulting risk approaches that of a clean-label oracle as $n$ grows. While the separation assumption is idealized, it approximates scenarios where corrupted examples remain high-loss across training. Our analysis deliberately remains at a high level due to space constraints, but the underlying argument closely follows existing self-paced learning theory~\cite{kumar2010selfpaced}: the alternating optimization over $(\theta,\gamma)$ induces a form of implicit curriculum where the optimal weight vector becomes sparse under suitable regularization. A more detailed treatment of convergence rates and finite-sample bounds for Step-E would require extending those results to loss-based hard gating and is left for future work.

\section{Experiments}
\subsection{Datasets}
\paragraph{CIFAR-100N.} CIFAR-100 contains 60{,}000 images from 100 classes (50{,}000 train, 10{,}000 test)~\cite{krizhevsky2009learning}. CIFAR-100N augments the training set with human-annotated noisy labels while providing both clean and noisy labels for each image~\cite{wei2022learning}. We use noisy labels for training and the clean test set for evaluation.

\paragraph{CIFAR-10N.} CIFAR-10 contains 60{,}000 images from 10 classes~\cite{krizhevsky2009learning}. CIFAR-10N provides multiple sets of human noisy labels; we use the ``aggre'' aggregation of crowd labels as the default noisy label set~\cite{wei2022learning} and the original clean test labels.

We deliberately focus on CIFAR-N with human-annotated noisy labels rather than synthetic label corruptions. As discussed in label-noise surveys~\cite{song2022survey,frenay2014classification}, simple corruption models such as class-independent flipping only coarsely approximate the structured, adversarial, or class-dependent noise patterns observed in practice. CIFAR-100N and CIFAR-10N provide a more realistic testbed where noise arises from human disagreements and annotation mistakes. Extending Step-E to a broader suite of synthetic corruption benchmarks and large-scale web-labeled datasets is an interesting direction for future work.

\subsection{Implementation details}
We use ResNet-18~\cite{he2016resnet} without pre-training for all experiments. Images are normalized by standard CIFAR means and standard deviations and augmented with random cropping and horizontal flipping. We optimize with SGD with Nesterov momentum $0.9$, weight decay $5\times10^{-4}$, batch size $128$, and cosine learning-rate decay starting from $0.1$. Mixed-precision training (AMP) runs on a single NVIDIA RTX 5090 GPU.

For both CIFAR-100N and CIFAR-10N we train for $60$ epochs with warm-up $T_{\mathrm{warm}}=10$, which is a standard budget for CIFAR-scale experiments and sufficient for all methods to reach stable test accuracy under our settings. Unless specified, Step-E uses $\rho_{\max}$ equal to the empirical noise rate plus a small margin and linearly ramps the drop ratio after warm-up. CIFAR-100N experiments are repeated with seeds $\{13,21,42\}$ and reported as mean $\pm$ standard deviation. CIFAR-10N results use a single seed ($42$) and appear in the appendix.

\subsection{Baselines}
All baselines share the same backbone and optimizer:
\begin{itemize}[leftmargin=1.5em]
    \item \textbf{Baseline.} Standard training on all noisy labels with cross-entropy loss and no cleaning.
    \item \textbf{Truncation.} Loss truncation where per-sample loss is clipped at a fixed threshold corresponding to the top $\rho_{\max}$ fraction of losses in an initial epoch.
    \item \textbf{Self-Paced.} Self-paced learning that gradually increases the keep ratio from a small value to $1$, eventually training on all samples.
    \item \textbf{One-Shot.} After warm-up, drop the top $\rho_{\max}$ high-loss samples once and then train on the remaining subset.
    \item \textbf{Oracle (clean).} Upper bound obtained by training the baseline model on clean CIFAR-100 labels.
\end{itemize}

\subsection{Main results on CIFAR-100N}
Table~\ref{tab:c100_main} reports test accuracy averaged over three seeds. Standard training reaches $43.3\%\,(\pm 0.7)$ under human label noise. Self-paced learning and one-shot filtering provide only marginal gains ($43.7\%\,(\pm 0.7)$ and $43.9\%\,(\pm 0.4)$, respectively), indicating that re-ordering or one-time pruning is insufficient. The truncation baseline collapses to $9.9\%\,(\pm 0.8)$, showing that static loss clipping can catastrophically suppress learning.

In contrast, Step-E attains $50.4\%\,(\pm 0.9)$, a $+7.1$-point improvement over the baseline. The clean-label oracle reaches $60.5\%\,(\pm 0.2)$.

\begin{table}[t]
    \centering
    \caption{Test accuracy on CIFAR-100N with human label noise (mean $\pm$ std over three seeds).}
    \label{tab:c100_main}
    \begin{tabular}{lc}
        \toprule
        Method & Test Acc. (\%) \\
        \midrule
        Baseline & $43.3\pm0.7$ \\
        Truncation & $9.9\pm0.8$ \\
        Self-Paced & $43.7\pm0.7$ \\
        One-Shot & $43.9\pm0.4$ \\
        Step-E (Ours) & \textbf{$50.4\pm0.9$} \\
        Oracle (clean labels) & $60.5\pm0.2$ \\
        \bottomrule
    \end{tabular}
\end{table}

\begin{figure}[t]
    \centering
    \includegraphics[width=0.8\linewidth]{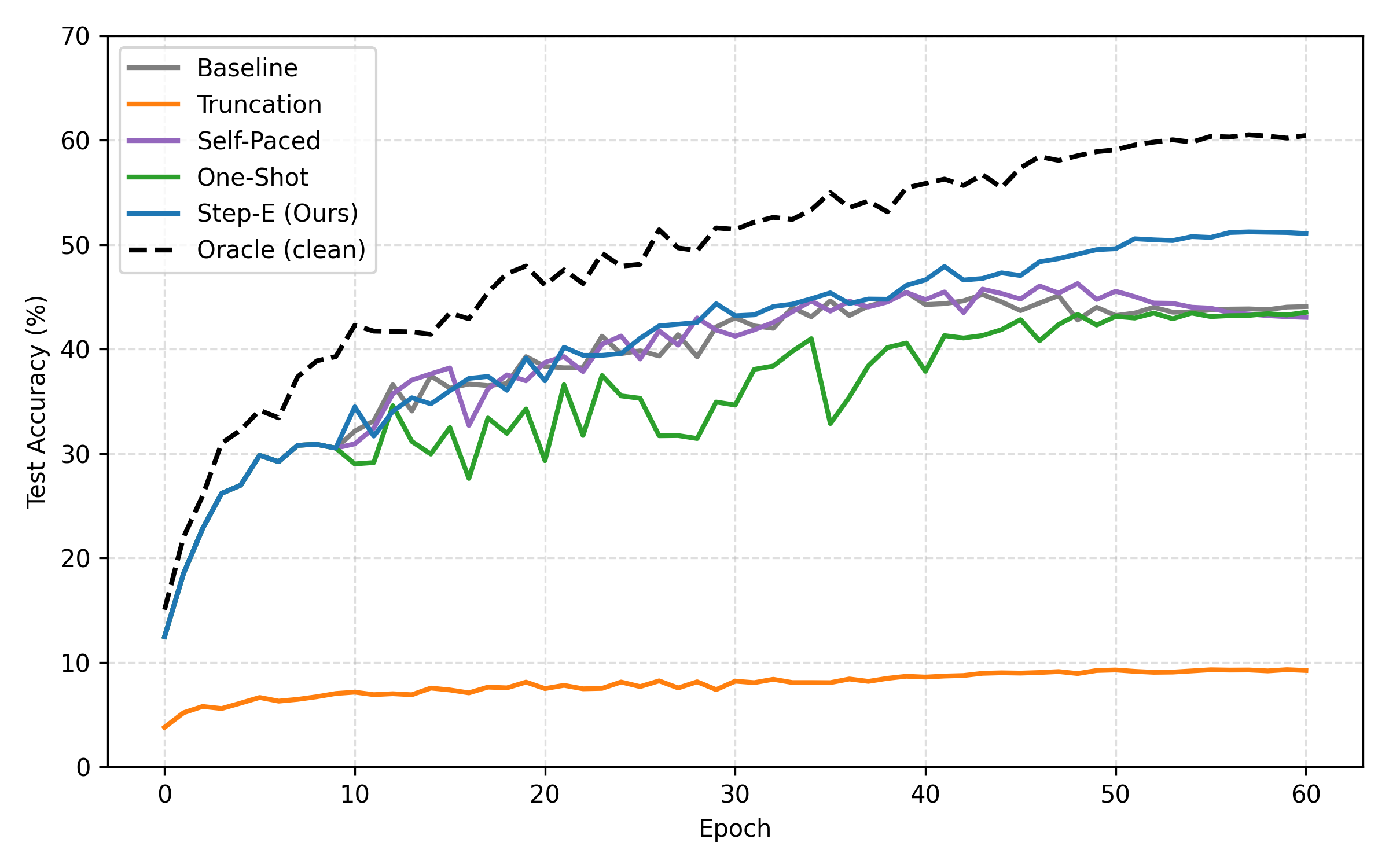}
    \caption{Test accuracy on CIFAR-100N across epochs (seed 42).}
    \label{fig:convergence_c100}
\end{figure}

Figure~\ref{fig:convergence_c100} plots test accuracy vs.\ epoch for all methods on CIFAR-100N (seed 42). Step-E converges to a higher final accuracy and shows stable mid-to-late training compared with the fluctuating one-shot baseline, while truncation remains near chance.

\subsection{Noise identification quality and training overhead}
Table~\ref{tab:noise_efficiency} reports noise detection quality and training-time overhead for methods that perform sample filtering. Noise detection is quantified with F1 and AUROC for predicting whether a training sample is noisy based on its loss ranking, measured at the final epoch using the current kept/dropped set and losses. Step-E improves both metrics to F1 $\approx72.2\%$ and AUROC $\approx84.8\%$, outperforming self-paced (F1 $=0.0$, AUROC $=78.7\%$) and one-shot (F1 $\approx62.4\%$, AUROC $\approx79.0\%$). Regarding efficiency, the baseline takes $13.85$ seconds per epoch on average; Step-E requires $19.17$ seconds, a $38.4\%$ overhead due to extra loss computation and sorting. Self-paced has similar overhead ($39.6\%$), and one-shot is slightly cheaper ($+20.7\%$).

\begin{table}[t]
    \centering
    \caption{Noise detection quality and training-time overhead on CIFAR-100N (mean over three seeds).}
    \label{tab:noise_efficiency}
    \begin{tabular}{lcccc}
        \toprule
        Method & Noise F1 (\%) & Noise AUROC (\%) & Time / epoch (sec) & Overhead (\%) \\
        \midrule
        Baseline & N/A & N/A & 13.85 & 0.0 \\
        Truncation & N/A & N/A & 13.82 & $-0.2$ \\
        Self-Paced & $0.0$ & $78.7$ & 19.34 & $39.6$ \\
        One-Shot & $62.4$ & $79.0$ & 16.72 & $20.7$ \\
        Step-E (Ours) & \textbf{$72.2$} & \textbf{$84.8$} & 19.17 & $38.4$ \\
        \bottomrule
    \end{tabular}
\end{table}

\subsection{Dynamics of Step-E}
\begin{figure}[t]
    \centering
    \includegraphics[width=\linewidth]{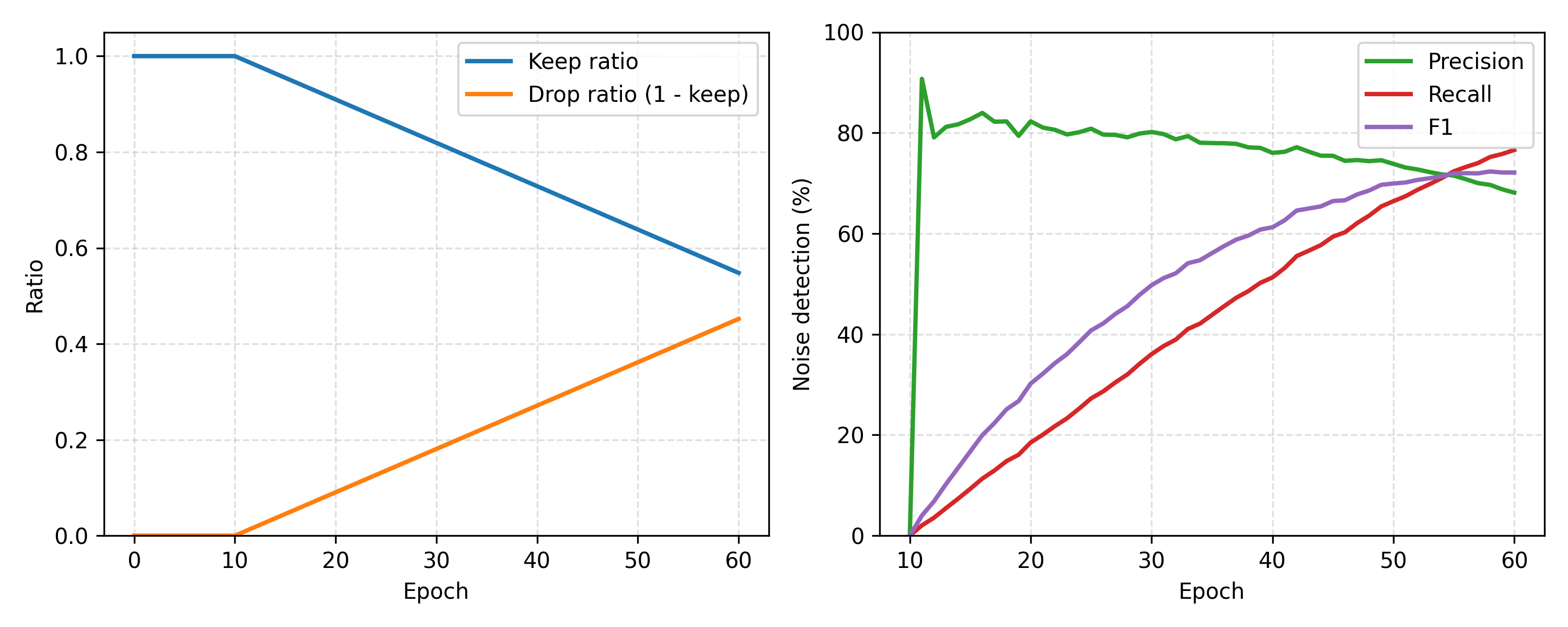}
    \caption{Dynamics of Step-E on CIFAR-100N (seed 42). Left: keep and drop ratios over epochs. Right: precision, recall, and F1 of noise detection.}
    \label{fig:dynamics}
\end{figure}

Figure~\ref{fig:dynamics} visualizes the evolution of keep/drop ratios and noise detection metrics on CIFAR-100N (seed 42). During warm-up (first 10 epochs), the keep ratio remains $1.0$ and the drop ratio $0$. After warm-up, the drop ratio increases linearly until reaching $\approx0.45$ at epoch 60, matching the estimated noise rate $\rho_{\max}$. Precision quickly rises to $80$--$90\%$ and then slowly decreases as recall grows. Recall increases nearly linearly to $\sim77\%$, and F1 stabilizes above $70\%$, indicating that Step-E progressively targets ambiguous high-loss examples while maintaining strong precision.

\subsection{Stability across random seeds}
\begin{figure}[t]
    \centering
    \includegraphics[width=0.8\linewidth]{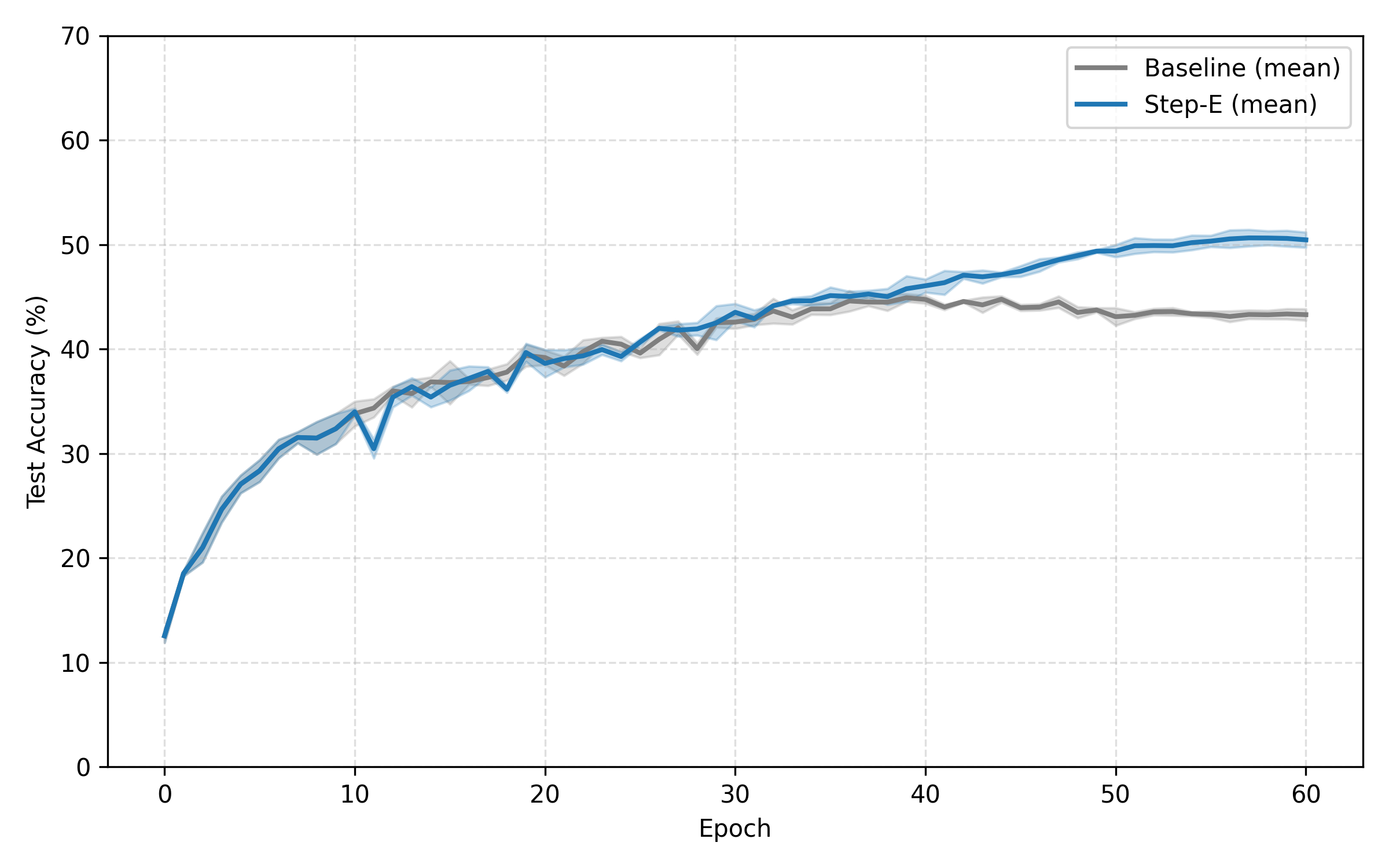}
    \caption{Mean $\pm$ standard deviation of test accuracy over three seeds on CIFAR-100N for baseline and Step-E.}
    \label{fig:seed_band}
\end{figure}

Figure~\ref{fig:seed_band} compares baseline and Step-E on CIFAR-100N using mean $\pm$ standard-deviation bands over three seeds. Step-E dominates the baseline throughout training with relatively narrow uncertainty bands. Table~\ref{tab:per_seed} in the appendix lists per-seed test accuracy and training time, confirming robustness to random initialization. Additional results on CIFAR-10N with human annotations are provided in Appendix~\ref{app:c10n}.

\section{Discussion and limitations}
Step-E provides substantial gains on CIFAR-N datasets with human label noise~\cite{wei2022learning} while remaining simple to implement and efficient to train. A large body of work studies learning from noisy labels~\cite{song2022survey,frenay2014classification}; our approach falls into the sample-selection family but focuses on integrating it tightly with SGD.

There are several limitations. First, we rely on the assumption that noisy examples maintain higher losses; systematic, easily learnable noise could evade loss-based filtering, suggesting value in adding uncertainty- or representation-based cues. Second, the warm-up length and linear ramp for $\rho_t$ are manually specified and could be learned or adapted automatically. Third, indiscriminate dropping may remove minority or rare patterns important for fairness or long-tail recognition, motivating fairness-aware or group-regularized extensions in the spirit of recent work on fairness in machine learning~\cite{mehrabi2021survey}.

\section{Conclusion}
Step-E tightly integrates sample selection with model training, dynamically removing high-loss samples via a stepwise elimination schedule and focusing learning on consistent examples without clean validation data or auxiliary models. On CIFAR-100N with human label noise, Step-E improves test accuracy by over seven percentage points over standard training and clearly outperforms loss truncation, self-paced learning, and one-shot filtering baselines while approaching the clean-label oracle. Similar gains appear on CIFAR-10N. Treating data cleaning as an integral component of optimization is a promising direction for building robust learning systems.

\bibliographystyle{plainnat}
\bibliography{refs}

\clearpage
\appendix
\section{Per-seed results on CIFAR-100N}
\label{app:per_seed}
Table~\ref{tab:per_seed} reports detailed per-seed results, including test accuracy, total training time, and noise detection metrics.

\begin{table}[h]
    \centering
    \small
    \caption{Per-seed results on CIFAR-100N. Test accuracy, total training time, and noise detection metrics for each method and seed.}
    \label{tab:per_seed}
    \begin{tabular}{lccccc}
        \toprule
        Method & Seed & Acc. (\%) & Time (s) & F1$_{\mathrm{noise}}$ (\%) & AUROC$_{\mathrm{noise}}$ (\%) \\
        \midrule
        Baseline & 13 & 42.9 & 855.4 & N/A & N/A \\
        Baseline & 21 & 42.9 & 841.3 & N/A & N/A \\
        Baseline & 42 & 44.1 & 837.4 & N/A & N/A \\
        Truncation & 13 & 10.8 & 839.5 & N/A & N/A \\
        Truncation & 21 & 9.8 & 844.1 & N/A & N/A \\
        Truncation & 42 & 9.2 & 846.1 & N/A & N/A \\
        Self-Paced & 13 & 44.4 & 1179.6 & 0.0 & 78.9 \\
        Self-Paced & 21 & 43.8 & 1179.8 & 0.0 & 78.5 \\
        Self-Paced & 42 & 43.0 & 1179.4 & 0.0 & 78.6 \\
        One-Shot & 13 & 43.8 & 1015.3 & 62.4 & 79.0 \\
        One-Shot & 21 & 44.4 & 1020.0 & 63.1 & 79.5 \\
        One-Shot & 42 & 43.5 & 1024.5 & 61.9 & 78.7 \\
        Step-E & 13 & \textbf{50.9} & 1170.7 & \textbf{72.5} & \textbf{85.0} \\
        Step-E & 21 & \textbf{49.4} & 1166.0 & \textbf{72.1} & \textbf{84.5} \\
        Step-E & 42 & \textbf{51.0} & 1171.0 & \textbf{72.1} & \textbf{84.9} \\
        Oracle (clean labels) & 13 & 60.7 & 844.5 & N/A & N/A \\
        Oracle (clean labels) & 21 & 60.3 & 845.9 & N/A & N/A \\
        Oracle (clean labels) & 42 & 60.4 & 844.5 & N/A & N/A \\
        \bottomrule
    \end{tabular}
\end{table}

\section{CIFAR-10N accuracy and curves}
\label{app:c10n}
While CIFAR-100N is the main benchmark due to its higher noise level and 100 classes, we also report results on CIFAR-10N (aggre) as a supplementary experiment. All methods train for 60 epochs with the same ResNet-18 backbone and hyperparameters (seed 42).

Because CIFAR-10N is a 10-class problem with relatively mild human label noise, the standard noisy-label baseline already attains high accuracy. In this near-saturated regime, absolute gains are naturally smaller, but Step-E still closes most of the remaining gap to the clean-label oracle: it improves the noisy baseline from $83.9\%$ to $85.3\%$, whereas the oracle reaches $85.9\%$ (Table~\ref{tab:c10_main}). This behavior is consistent with our findings on CIFAR-100N, where the larger label space and higher noise rate make the benefits of data cleaning more pronounced.

\begin{figure}[t]
    \centering
    \includegraphics[width=0.7\linewidth]{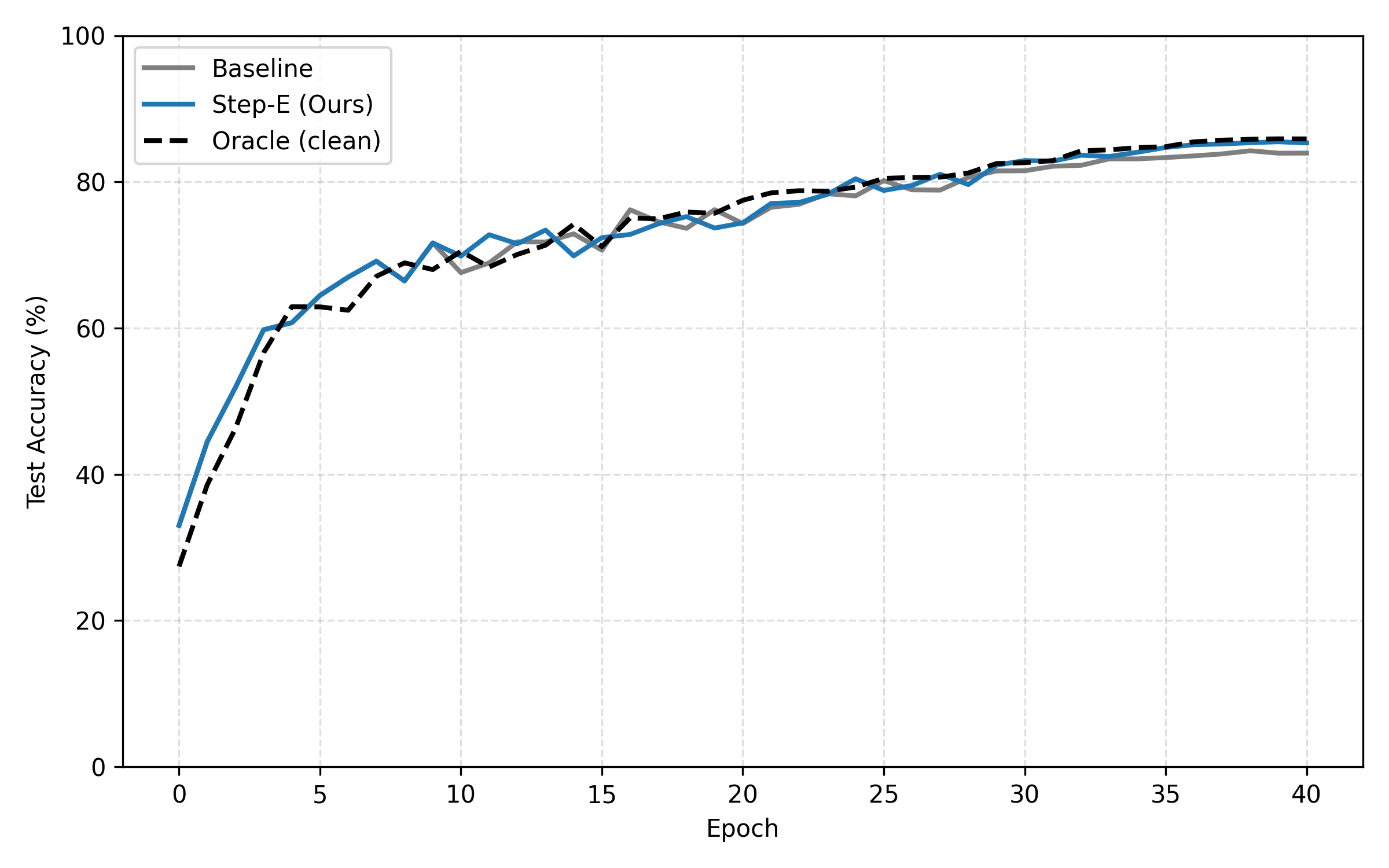}
    \caption{Test accuracy on CIFAR-10N (aggre) vs.\ epoch for baseline, Step-E, and clean-label oracle (seed 42).}
    \label{fig:c10_curves}
\end{figure}

\begin{table}[t]
    \centering
    \caption{Test accuracy (\%) on CIFAR-10N (aggre) for a single run (seed 42).}
    \label{tab:c10_main}
    \begin{tabular}{lc}
        \toprule
        Method & Test Acc. (\%) \\
        \midrule
        Baseline & 83.9 \\
        Step-E (Ours) & \textbf{85.3} \\
        Oracle (clean labels) & 85.9 \\
        \bottomrule
    \end{tabular}
\end{table}

\end{document}